\pdfoutput=1
\documentclass[11pt]{article}

\usepackage[]{acl}

\usepackage{times}
\usepackage{latexsym}

\usepackage[T1]{fontenc}
\usepackage[utf8]{inputenc}

\usepackage{microtype}

\usepackage{enumitem}
\setitemize{noitemsep,topsep=0pt,parsep=0pt,partopsep=0pt}
\definecolor{cite}{rgb}{0.6,0.6,1.0}

\definecolor{todo}{rgb}{1,0.5,0}

\usepackage{booktabs}
\usepackage{graphicx}
\usepackage{float}
\usepackage{tabularx}
\usepackage{soul}
\usepackage{subcaption}
\usepackage{multicol}
\usepackage{multirow}
\usepackage{amsfonts}
\usepackage{tabularx}
\usepackage{makecell}

\title{Document Structure in Long-Document Transformers}

\author{Jan Buchmann$^{1*}$, Max Eichler$^{1*}$, Jan-Micha Bodensohn$^{2}$, Ilia Kuznetsov$^1$, Iryna Gurevych$^1$ \\
  $^1$ Ubiquitous Knowledge Processing Lab (UKP Lab) \\
    Department of Computer Science and Hessian Center for AI (hessian.AI)\\
    Technical University of Darmstadt \\
    \url{www.ukp.tu-darmstadt.de} \\
   $^2$ DFKI and Data and AI Systems Lab, 
Technical University of Darmstadt \\
}

\begin{document}
\maketitle

\def\thefootnote{*}\footnotetext{Equal contribution}\def\thefootnote{\arabic{footnote}}

\begin{abstract}
Long documents often exhibit structure with hierarchically organized elements of different functions, such as section headers and paragraphs. Despite the omnipresence of document structure, its role in natural language processing (NLP) remains opaque. Do long-document Transformer models acquire an internal representation of document structure during pre-training? How can structural information be communicated to a model after pre-training, and how does it influence downstream performance? To answer these questions, we develop a novel suite of probing tasks to assess structure-awareness of long-document Transformers, propose general-purpose structure infusion methods, and evaluate the effects of structure infusion on QASPER and Evidence Inference, two challenging long-document NLP tasks. Results on LED and LongT5 suggest that they acquire implicit understanding of document structure during pre-training, which can be further enhanced by structure infusion, leading to improved end-task performance. To foster research on the role of document structure in NLP modeling, we make our data and code publicly available\footnote{\url{https://github.com/UKPLab/eacl2024-doc-structure}, under Apache-2.0 license.}.

\end{abstract}

\section{Introduction}

\begin{figure}[ht]
    \centering
    \includegraphics{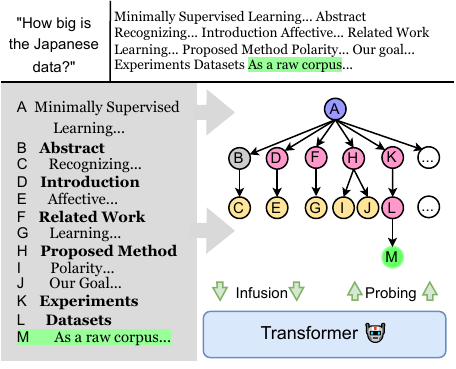}
    \caption{Transformer models receive unstructured text as input (top right) -- yet long texts exhibit structure, which helps in finding information (bottom). We investigate whether Transformers learn representations of document structure during pre-training (§\ref{sec:probing_for_structure}), whether structure-awareness can be enhanced by infusion after pre-training (§\ref{sec:structure_infusion}), and what effects infusion has on downstream task performance. Source: QASPER dataset, arxiv ID 1909.00694 \cite{dasigi-etal-2021-dataset}.}
    \label{fig:eyecatcher}
\end{figure}

Long documents such as news articles, scientific papers, and clinical reports play a vital role in many human activities. 
These documents are usually organized into chapters, sections, subsections, and paragraphs, i.e. they are structured. This helps humans in navigating documents \cite{guthrie_roles_1991, nguyen-etal-2021-skim-attention} and building a mental model of the content \cite{taylor_effects_1984, meyer_use_1980}. The example in Fig.~\ref{fig:eyecatcher} shows how the hierarchy of sections and subsections helps when looking for the size of a dataset in an NLP paper: one would go via the "Experiments" section to the "Datasets" subsection.

Although structure is omnipresent and useful to humans, existing long-document Transformers (e.g. \citealt{ainslie-etal-2020-etc, beltagy2020longformer, 10.1162/tacl_a_00547}) operate with linearized textual input: documents are converted to flat character strings, removing the distinction between different functional elements and their hierarchy (Fig.~\ref{fig:eyecatcher}, top right). 

Understanding the structural capabilities of long-document Transformers is important both theoretically and practically. From a theoretical standpoint, prior work in probing has demonstrated the ability of Transformers to learn syntactic representations on the sentence level \cite{hewitt-liang-2019-designing} -- yet little is known about the ability to induce higher-level discourse structures from linearized text. Probing methodology and datasets for this investigation are missing. From a practical perspective, recent works demonstrate that structure-aware modeling can improve downstream task performance \cite{li2023towards, cao-wang-2022-hibrids, ruan-etal-2022-histruct} -- yet existing studies are limited to task-specific architectures and data formats, making it hard to generalize the findings to new tasks and document types. General-purpose methodology for communicating structural information to Transformer models is yet to be established.

Our work aims to close this gap. Instead of committing to a specific document format, we build the a task- and format-agnostic formalism of Intertextual graphs (ITG, \citealt{10.1162/coli_a_00455}) to encode structure obtained from the original documents (§\ref{sec:formalizing_structure}).

Building on this formalism, we investigate the role of document structure in long document Transformers from two experimental angles: Probing and downstream tasks. We introduce a novel suite of probing tasks in §\ref{sec:probing_for_structure} to investigate structure-awareness of pre-trained Transformer models. Probing experiments on two widely used long document Transformer models -- LED \cite{beltagy2020longformer} and LongT5 \cite{guo-etal-2022-longt5} -- suggest that Transformers do acquire the ability to represent document structure during pre-training, but that there is room for improvement. Consequently, in §\ref{sec:structure_infusion}, we test the effect of adding structural information to the Transformer input. We devise a general-purpose structure infusion kit and employ it in experiments on our probing suite and two challenging long-document NLP datasets: QASPER \cite{dasigi-etal-2021-dataset} and Evidence Inference \cite{deyoung-etal-2020-evidence}. The results suggest that structure-awareness can be enhanced via infusion, leading to up to 6.8 F1 points increase on downstream tasks. Our work lays the foundation for the systematic analysis of the role of document structure in long document modeling.

\section{Background}
\label{sec:background}

\begin{figure}[ht]
    \centering
    \includegraphics{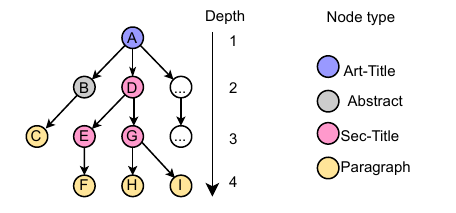}
    \caption{Document Graph. Black arrows show \texttt{parent} edges, \texttt{next} edges between alphabetically consecutive nodes are omitted for clarity. Node depth and node type information are infused in §\ref{sec:structure_infusion}.}
    \label{fig:graph}
\end{figure}

\paragraph{Document structure.} 
The term "structure" is used ambiguously for textual documents. \textit{Rhetorical structure} is the hierarchical organization of semantic units, usually latent and not available for explicit processing. \cite{kintsch-van-dijk-toward, mann1987rhetorical}. \textit{Abstract structure} refers to the hierarchical organization of a text into elements such as sections, paragraphs, and lists\footnote{Power et al. (\citeyear{power_document_2003}) include phenomena such as emphasis and quotation into abstract document structure. They are not considered here, as they are rarely preserved or standardized.} \cite{nunberg1990linguistics, power_document_2003}. \textit{Concrete}, or \textit{visual structure}, includes aspects of typesetting such as font size, spacing and the location of textual elements in a typeset text, classically ordered into pages \cite{power_document_2003}. In this work, we focus on the study of abstract document structure  as the direct author expression of textual organization. 

\paragraph{Long-document Transformers.}
The memory and computational requirements of the standard Transformer architecture \cite{NIPS2017_3f5ee243} scale quadratically with the input length, making it hard to process long documents under computational constraints. Several innovations for increased efficiency have been proposed, surveyed by Tay et al. (\citeyear{10.1145/3530811}). A popular and well-performing approach is the combination of local attention with a varied distribution of global attention \cite{ainslie-etal-2020-etc, beltagy2020longformer, guo-etal-2022-longt5}, used by the top 5 models in the Scrolls benchmark for long-document processing \cite{shaham-etal-2022-scrolls}. We experiment with two representatives for this approach: LED \cite{beltagy2020longformer}, which is employed in many recent works on long documents (e.g. \citealt{dasigi-etal-2021-dataset, cao-wang-2022-hibrids}) and LongT5 \cite{guo-etal-2022-longt5}, the best "base" model on the Scrolls leaderboard at the time of writing\footnote{\url{https://www.scrolls-benchmark.com/leaderboard}, October 2023.}. 

\paragraph{Probing.} Probing tasks are diagnostic classification tasks which investigate whether a linguistic feature (e.g. sentence length, word content or syntax tree depth) is encoded in a representation \cite{conneau-etal-2018-cram, belinkov-2022-probing, rogers-etal-2020-primer}. Early work on probing measured the encoded knowledge through the delta to a majority baseline or randomly initialized embeddings. Control tasks were introduced as a better approximation of what a probing classifier is able to learn in its own neural representation compared to what linguistic features it can extract from the underlying representations~\cite{hewitt-liang-2019-designing}.
We follow this line of work by designing a novel atomic control setting where we remove contextual information. To measure contextual information beyond a given span, we employ edge probing introduced by Tenney et al., (\citeyear{tenney-etal-2019-bert}).

Syntax trees have been shown to be encoded in BERT~\cite{hewitt-manning-2019-structural}, but the representation of higher-order document structure has not been investigated. For the first time, we show that long-document Transformers internally represent several aspects of document structure, and that this internal representation can be enhanced.

\paragraph{Document structure in Transformers.} 
Existing approaches that make use of abstract document structure in Transformers broadly fall into two categories. In \textit{hierarchical processing} \cite{zhang-etal-2022-hegel, qi-etal-2022-sapgraph, liu-lapata-2019-hierarchical, ruan-etal-2022-histruct}, complex, task specific architectures are built, from which results and analyses are hard to generalize. In \textit{structure infusion}, additional structural information is added to pre-trained Transformer models. We employ the latter setting, because methods and models can be reused and analyzed more easily. Structure infusion through special tokens (\citealt{aghajanyan2022htlm, fisch-etal-2019-mrqa}), attention masks \cite{liu-etal-2021-hetformer, hong-etal-2022-graph}, absolute (\citealt{Bai_Shi_Lin_Xie_Tan_Xiong_Gao_Li_2021}) or relative position embeddings \cite{cao-wang-2022-hibrids} has been shown to improve downstream task performance. Here, we combine special tokens and position embeddings which only require changes at the input layer, making them easily transferrable to other transformer models.

\section{Representing Structure}
\label{sec:formalizing_structure}

\paragraph{Formalism.} We model the abstract structure of a document (\citealt{power_document_2003}, see §\ref{sec:background}) as an ordered graph $G$ (Fig.~\ref{fig:graph}) as in Kuznetsov et al.~(\citeyear{10.1162/coli_a_00455}), using their notation. Structural elements such as section headings or paragraphs are represented as a set of typed nodes $N^G$. The node types correspond to the \textit{function} of the element in the document. We consider the types \texttt{article-title}, \texttt{section-title}, \texttt{abstract}, and \texttt{paragraph}\footnote{We do not consider sentences, as their borders often cannot be extracted unambiguously from English texts.}. The set of typed, directed edges $E^G$ encodes the \textit{hierarchical organization} of the textual elements with \texttt{parent} edges and the linear order with \texttt{next} edges. Node function and hierarchical organization can be seen as orthogonal pieces of information that together fully describe the abstract document structure.

\paragraph{Data conversion.} All datasets used in the present work were converted to the intertextual graph (ITG) format\footnote{\url{https://github.com/UKPLab/intertext-graph}} introduced in Kuznetsov et al.~(\citeyear{10.1162/coli_a_00455}), which is a generic JSON representation of the graph data structure introduced above. Many different types of documents can be easily converted to the ITG format without loss of information on the document structure, including XML or \LaTeX files. All our methods and experiments are based on ITG, and are therefore dataset agnostic, easily adaptable, and extensible. 

\section{Probing for Structure}
\label{sec:probing_for_structure}

\subsection{Probing Suite Design}
\label{sec:probing_for_structure/probing_tasks}

\begin{table*}
    \centering
    \noindent\begin{tabularx}{\linewidth}{l|X|l}
    \toprule
    Name & Classification task & Labels \\
    \midrule
    \texttt{Node type} & Type of $n_j$ with all nodes of type \texttt{section} and a tree depth $> 1$ grouped as \texttt{subsection}[1]. & \makecell{\texttt{Section}, \\ \texttt{subsection}, \\ \texttt{paragraph}} \\
    \texttt{Sibling} & Do $n_j$ and $n_k$ share the same parent $n_p$? & \texttt{Boolean} \\
    \texttt{Ancestor} & Is $n_j$ on the \texttt{parent} path of $n_k$ and the root $n_0$? & \texttt{Boolean} \\
    \texttt{Position} & Position within an ordered set $S$ for all nodes $n_j \in S$ with the same parent $n_p$. & \makecell{\texttt{Begin}, \texttt{inside}, \\ \texttt{outside}} \\
    \texttt{Parent predecessor} & Is $n_p$ the parent of $n_j$? & \texttt{Boolean} \\
    \texttt{Tree depth} & Depth of $n_j$ from the root $n_0$. & \texttt{Integer} \\
    \texttt{Structural} & Shortest \texttt{parent} path between $n_j$ and $n_k$. & \texttt{Integer} \\
    \bottomrule
    \end{tabularx}
    \caption{Definitions of probing tasks and their labels. $n_{j,k,p,0}$ denote nodes in the document graph $G$. [1] \texttt{Subsection} is a mixture of functional and hierarchical description, so it is not part of the node types defined in §\ref{sec:formalizing_structure}. It is added to the \texttt{node type} probing task to increase the difficulty.}
    \label{tab:probing_tasks}
\end{table*}

As the first step towards the systematic study of document structure in long document processing, we propose a suite of seven probing tasks that measure the ability of pre-trained Transformers to capture structural information from their input, described in Tab.~\ref{tab:probing_tasks}. For example, the \texttt{parent predecessor} probe measures the representation of document hierarchy in a Transformer by learning to distinguish between pairs of document elements (e.g. headings or paragraphs) that are in a parent-child relationship and pairs that are not. As shown in our introduction example, a good representation of the hierarchy can help in locating relevant information in a document (Fig.~\ref{fig:eyecatcher}).

All probing tasks are cast as classification and evaluated via accuracy. Assuming a model that computes vector representations of textual nodes, classification is implemented as a linear layer projecting from the representation of a node or a node pair to the label space. If a model has multiple layers, node representations are computed as a weighted sum ~\cite{tenney-etal-2019-bert} of the representations from each layer. For tasks on node pairs, the representations of two nodes are concatenated. Only the linear layer and the scalar mix weights are updated during training on the probing task. 

\subsection{Experiments and Results}
\label{sec:probing_for_structure/results}

\paragraph{Probing dataset.} We instantiate our probing tasks with research papers from the open science platform F1000Research\footnote{\url{https://F1000research.com}, downloaded on April 9th, 2021. We use the paper first versions.}. Based on the pre-processing used for the F1000RD corpus \cite{10.1162/coli_a_00455} we convert each paper into the ITG format (Fig. \ref{fig:graph}), removing all non-textual nodes\footnote{For the node type probe we remove the document title and abstract as well, as these occur once per document.}. Removing all papers exceeding the maximum input length of LED (16384 tokens) results in a corpus of 2,499 documents. All probing tasks are balanced through downsampling on document basis, meaning that the label distribution is uniform in most cases (Tab.~\ref{tab:dataset}). For some probes, e.g. \texttt{tree depth}, not all labels occur in all documents, resulting in a non-uniform label distribution.

\begin{figure*}
    \centering
    \includegraphics{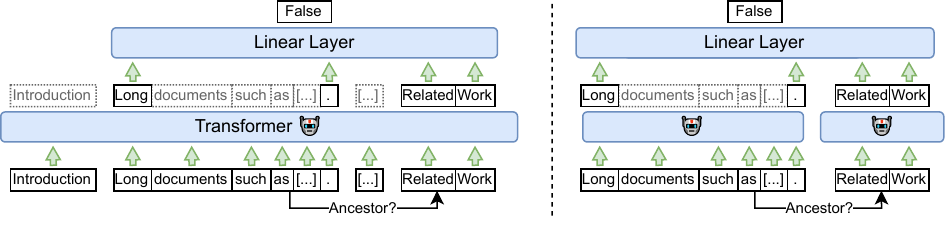}
    \caption{Probing classifier with the vanilla probing architecture encoding a full document (left) and the atomic encoding two nodes individually without any context (right). Tokens w/ arrow are used as input to the next layer.}
    \label{fig:probing_architecture}
\end{figure*}

\paragraph{Probing architecture.} We compare probing of the "vanilla" LED and LongT5 encoders with two control configurations each: \emph{atomic} and \emph{random}. In the atomic control (Fig.~\ref{fig:probing_architecture}), nodes are input to the model individually, i.e. without their document context. 
Comparing the vanilla and atomic configurations shows the effect of contextualization on the representation of structure. For the random control, all model weights except for the embedding layer are re-initialized randomly~\cite{jawahar-etal-2019-bert}. It shows the effect of pre-training on the representation of structure. Details on implementation and hyperparameters can be found in Appx.~\ref{sec:appendix/implementation/probing}.

\paragraph{Results.} In all probes, the accuracy of the vanilla model is higher than the random control (Tab.~\ref{tab:probing_baselines}). The difference varies between $34\%$ for LongT5 on \texttt{position} and $2.7\%$ for LED on \texttt{node type} -- a magnitude comparable to reported results from prior work on probing (e.g. \citealt{conia-navigli-2022-probing}). This result suggests that LED and LongT5 learn to represent document structure during pre-training, but the effect varies between different aspects of document structure. The cases with small difference between vanilla and random control imply that the input token and position embeddings, not being re-initialized, contain much of the information needed to solve the task. The scores of the atomic control are lower than those of the vanilla configuration on all probes, showing that context helps to represent document structure.

Vanilla LED and LongT5 achieve accuracies of $~$0.9 on some probes, e.g. \texttt{node type}, suggesting that they are able to encode some aspects of structural information well even without its explicit input. It is surprising that the accuracy on the \texttt{sibling probe} is far below that of \texttt{parent predecessor}, because the information on the parents of two nodes is enough to determine their siblinghood. It seems that the combination of parent information from two nodes in a queried pair is difficult. The \texttt{structural} probe can be considered the most complex, as it has the most classes. Thus, the large room for improvement is expected.

We could show for the first time that long-document Transformers can learn to represent document structure, even though the models were not explicitly trained for this. However, the representation of some aspects of structure is far from optimal. In the following, we investigate whether structure infusion, i.e. the input of additional, explicit information on document structure, improves the internal representation of structure and if this translates to improvements on downstream tasks. 

\begin{figure*}
    \centering
    \includegraphics{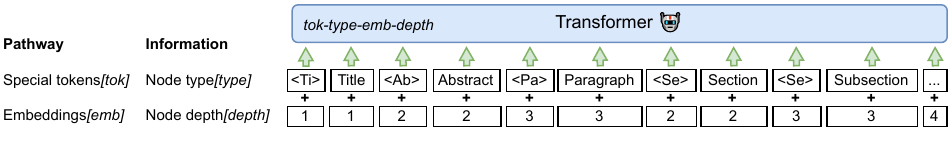}
    \caption{Structure infusion via special tokens and embeddings. Special tokens ("<Ti>", "<Ab>") are prepended to the text of the corresponding node, embeddings are summed with the token embeddings. The figure shows the combination of hierarchical embeddings and node type special tokens, short description \texttt{tok-type-emb-depth}.}
    \label{fig:structure_infusion}
\end{figure*}

\section{Infusing Structure}
\label{sec:structure_infusion}

As exemplified in Fig.~\ref{fig:eyecatcher}, abstract document structure can help humans in working with documents. While previous work shows that the addition of structural information can improve the downstream performance of Transformer models \cite{li2023towards, cao-wang-2022-hibrids, ruan-etal-2022-histruct}, the use of task-specific architectures and document formats prevents comparison of structure infusion methods across the studies, and makes it challenging to relate performance to probing results. To remedy this, we introduce a task- and format-agnostic structure infusion kit, and demonstrate its wide applicability by studying the effects of structure infusion on LED and LongT5 and two challenging long-document tasks.

\subsection{Methodology~\protect\footnote{We provide implementation details in Appx.~\ref{sec:appendix/implementation/infusion}-\ref{sec:appendix/implementation/computation}.}}
\label{sec:structure_infusion/methodology}

\begin{table*}[]
    \centering
    \begin{tabular}{lrrrrrrr}
\toprule
  & Nod & Sib & Anc & Pos & Par & Tre & Str \\
\midrule
LED & \textbf{93.98} & \textbf{64.93} & \textbf{89.53} & \textbf{86.05} & \textbf{85.68} & \textbf{84.12} & \textbf{41.49} \\
LED Atom & \underline{92.75} & \underline{60.26} & \underline{87.30} & \underline{65.53} & \underline{84.82} & \underline{82.41} & \underline{40.64} \\
LED Rand & 88.21 & 58.36 & 86.73 & 56.44 & 82.90 & 73.76 & 35.33 \\
\midrule
LongT5 & \textbf{95.28} & \textbf{65.85} & \textbf{89.38} & \textbf{91.95} & \textbf{86.13} & \textbf{87.88} & \textbf{42.97} \\
LongT5 Atom & \underline{91.84} & 50.79 & \underline{86.60} & \underline{61.05} & \underline{83.77} & \underline{78.90} & \underline{34.68} \\
LongT5 Rand & 88.21 & \underline{57.41} & 84.81 & 57.97 & 81.54 & 73.40 & 33.49 \\
\bottomrule
\end{tabular}
    \caption{Probing accuracy of LED and LongT5 with atomic and random controls. Best result per model and probe in bold, second best underlined.}
    \label{tab:probing_baselines}
\end{table*}

\begin{figure*}
    \centering
    \includegraphics{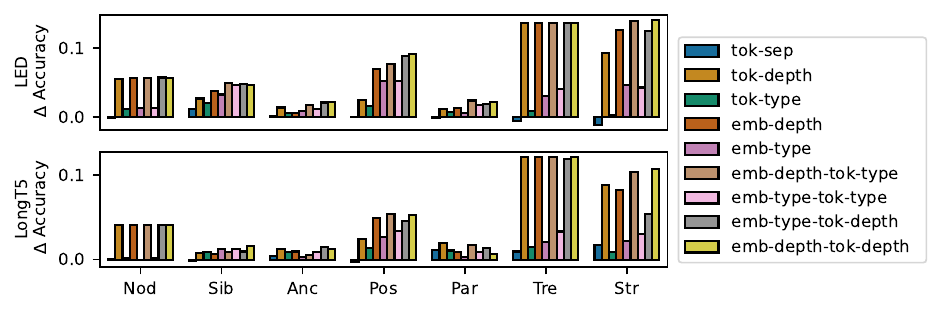}
    \caption{Probing of structure-infused models. Bars show the difference in accuracy to the vanilla baseline (Tab.~\ref{tab:probing_baselines}) For absolute values see Tab.~\ref{tab:probing}.}
    \label{fig:probing_structure_infusion}
\end{figure*}

\begin{table*}
    \centering
    \begin{tabular}{lrr|rr||rr|rr}
\toprule
{} & \multicolumn{4}{c||}{LED} & \multicolumn{4}{c}{LongT5} \\
{} & \multicolumn{2}{c|}{QAS} & \multicolumn{2}{c||}{EvI} & \multicolumn{2}{c|}{QAS} & \multicolumn{2}{c}{EvI} \\
{} &   Ans &   Evi &   Cla &   Evi &    Ans &   Evi &   Cla &   Evi \\
\midrule
vanilla              & 36.80 & 42.05 & 74.30 & 61.55 &  45.89 & 52.09 & \textbf{81.54} & 70.39 \\
\midrule
tok-sep       & 37.35 & 42.54 & 75.17 & 66.81 &  45.54 & 54.12 & 81.08 & 75.92 \\
tok-depth           & 36.24 & 41.90 & 74.60 & 64.19 &  \underline{46.60} & \textbf{56.14} & 80.90 & \underline{76.88} \\
tok-type            & 37.43 & 42.32 & 75.85 & \underline{66.93} &  \textbf{46.76} & \underline{56.08} & 80.75 & 76.28 \\
emb-depth          & 36.17 & 42.53 & 73.78 & 60.67 &  44.91 & 51.53 & 81.36 & 71.18 \\
emb-type            & 36.03 & 42.92 & 74.71 & 61.05 &  46.37 & 53.89 & 80.86 & 68.91 \\
emb-depth-tok-type & 37.83 & 43.16 & \textbf{76.49} & 66.07 &  45.63 & 56.04 & 79.94 & 75.57 \\
emb-type-tok-type  & \underline{38.02} & 43.83 & \underline{76.38} & 65.31 &  46.43 & 55.70 & \underline{81.42} & \textbf{77.23} \\
emb-type-tok-depth  & \textbf{39.08} & \underline{44.41} & 75.30 & 64.58 &  44.72 & 55.60 & 80.71 & 75.86 \\
emb-depth-tok-depth & 37.74 & \textbf{44.64} & 76.34 & \textbf{67.07} &  45.33 & 54.27 & 80.98 & 75.96 \\
\bottomrule
\end{tabular}
    \caption{Downstream task results on test sets. All scores are F1 scores averaged over 3 runs with different random seeds. Best result in column in bold, second best underlined. QAS: QASPER. EvI: Evidence Inference. Ans: Answer F1. Evi: Evidence F1. Cla: Classification F1.}
    \label{tab:downstream_tasks}
\end{table*}

\paragraph{Structure infusion.} We infuse information on abstract document structure through position embeddings added to the token embeddings (indicated as \texttt{emb}, see Fig.~\ref{fig:structure_infusion}) and special tokens that are prepended to the tokens of the corresponding node (\texttt{tok}). Both methods only modify the input layer and are therefore easily applicable to any Transformer model.

We infuse the two types of abstract structural information that are missing in the input of Transformer models (§\ref{sec:formalizing_structure}): node function and hierarchy. Node function is infused through embeddings and special tokens representing the node type (\texttt{type}). To infuse the hierarchical organization, tokens and embeddings represent the depth of a node in the graph, i.e. its distance to the document root (\texttt{depth}). As a baseline for structural tokens, we prepend each node with the same separator token (\texttt{sep}). We refer to the infusion configurations using short descriptors, e.g. the combination of node depth position embeddings and node type tokens is shortened to \texttt{emb-depth-tok-type}. 

\paragraph{Probing.} The probing experiments were conducted as described in §\ref{sec:probing_for_structure} using the same probing dataset, with the addition of structural information in the input. We omit the atomic and random control here, as we are interested in the capabilities of the configuration that is used for downstream tasks.

\paragraph{Downstream task datasets.} We selected QASPER \cite{dasigi-etal-2021-dataset} and Evidence Inference \cite{deyoung-etal-2020-evidence} by the following criteria: they are based on long documents, abstract document structure is available, and several types of downstream tasks are covered, to see possible differences in the effect of structure infusion.

\textbf{QASPER} is a collection of scientific papers from computational linguistics / NLP and corresponding questions with one or multiple answers with evidence. We model question answering as a generative problem and evidence selection as paragraph classification. Answer generation and evidence selection are evaluated with F1 scores using the evaluation script provided by the authors\footnote{https://github.com/allenai/qasper-led-baseline}.

\textbf{Evidence Inference} consists of reports from clinical studies, "prompts" in the form of \textit{intervention}, \textit{comparator}, and \textit{outcome}, one or multiple labels for the prompt ("significantly increased", "significantly decreased", or "no significant difference") and corresponding evidence spans. We model prompt answering as 3-way classification, and convert evidence span selection to node classification by mapping evidence spans to nodes. As there is no adaptable evaluation script, and for consistency with QASPER, we re-implemented evaluation, choosing the annotation resulting in the highest score as gold standard. This means that we can only compare the models in our work.

\paragraph{Training} Downstream tasks were fine-tuned for 10,200 steps with an effective batch size of 8 in a multi task fashion. We report mean test set results of 3 random seeds.

In all experiments in this section, the models were pre-trained for 15,000 steps, with an effective batch size of 16, with the respective structure infusion configuration on the relevant probing (F1000RD) or downstream task dataset (QASPER or Evidence Inference), as we noted this to be beneficial in early experiments \cite{gururangan-etal-2020-dont}. "T5-style" denoising \cite{10.5555/3455716.3455856} was used as the pre-training task as suggested in Xiong et al, (\citeyear{xiong2022adapting}).

\subsection{Probing of Structure-Infused Models}
\label{sec:structure_infusion/probing_with_structure}

We see an improvement in all probes through structure infusion (Fig.~\ref{fig:probing_structure_infusion}, Tab.~\ref{tab:probing}). The \texttt{node type} and \texttt{tree depth} probes show an accuracy of around 1 with tree depth infusion, as this information suffices to solve the tasks. Node type infusion does not lead to perfect scores on the \texttt{node type} probe, as the subsection node type is part of the probing task, but not of the infusion (Tab.~\ref{tab:probing_tasks}).

Except for LongT5 on \texttt{sibling}, infusion of node depth results in higher accuracy than node type or node boundary information infused on the same pathway. For the majority of LED probes (\texttt{sibling}, \texttt{position}, \texttt{tree depth}, and \texttt{structural}), models with position embedding infusion show higher metrics than their counterparts with the same information in special tokens, while for LongT5, the results are mixed. LED, based on BART \cite{lewis-etal-2020-bart}, is pre-trained with absolute position embeddings like our structural embeddings, while LongT5, based on T5 \cite{10.5555/3455716.3455856}, uses relative position embeddings. LED might therefore have a better capability to use the information from absolute embeddings. 

\subsection{Structure infusion in Downstream Tasks}
\label{sec:structure_infusion/downstream_tasks}

\paragraph{QASPER} For LED in answer generation, the \texttt{emb-type-tok-depth} configuration results in the best performance, with an improvement of 2.28 F1 points over vanilla (Tab.~\ref{tab:downstream_tasks}). In evidence selection, \texttt{emb-depth-tok-depth} outperforms the vanilla configuration by 2.59 F1 points. This is an improvement of 5.58 F1 points for answer generation and 14.04 F1 points for evidence selection over the LED state-of-the-art (SOTA) \cite{caciularu-etal-2022-long} on QASPER. The vanilla configuration already outperforms the SOTA by 3.30 and 11.45 F1 points, respectively. Infusing the node depth through two pathways improves over a single pathway. While unintuitive, this was also observed for the \texttt{sibling}, \texttt{parent predecessor}, and \texttt{tree depth} probes (Fig.~\ref{fig:probing_structure_infusion}).

For LongT5, special tokens structure infusion results in the highest scores. The best answer F1 of 46.76 with node type tokens improves the vanilla model by 0.87 points and is slightly higher than the current LongT5-base SOTA of 46.6 \cite{guo-etal-2022-longt5}. In evidence selection, infusion of depth tokens increases the vanilla configuration by 4.05 F1 points. To our knowledge, there are no reported scores for LongT5 on QASPER evidence selection. 

\paragraph{Evidence Inference} For LED, the best performance in classification is obtained by the \texttt{emb-depth-tok-type} configuration, improving $2.19$ F1 points over the vanilla configuration. In evidence selection, \texttt{emb-depth-tok-depth} outperforms the vanilla baseline by 5.52 F1 points, but adding node separator tokens already leads to an increase of 5.26 F1 points. 

For LongT5, no structure infused variant outperforms vanilla in classification, while in evidence selection, \texttt{emb-type-}\texttt{tok-type} outperforms vanilla by 6.84 F1 points.

\paragraph{Comparison of infusion configurations.} In most cases, adding node separator tokens improves performance. This was expected, as it is common practice to signify segment boundaries to models (e.g. \citealt{beltagy2020longformer}) and could also be seen in probing. For LED, the combination of position embeddings and structural tokens exhibits the best scores, which again resembles the probing results. For LongT5, combining both infusion pathways only results in the best scores on Evidence Inference evidence selection. Infusion via structural tokens outperforms infusion via position embeddings for LongT5 on most subtasks. 

The increases for LED of about 2 F1 points are similar to the reported performance increases through document structure infusion on other long-document datasets, showing that our employed methods are effective. These works use relative position embeddings \cite{cao-wang-2022-hibrids} or special attention patterns \cite{liu-etal-2021-hetformer, hong-etal-2022-graph}, while we use structural tokens and absolute position embeddings. Our methods are easier to apply and adapt, as only the input to the model needs to be modified. For LongT5, the performance gains through structure infusion of up to 6.84 F1 points suggest that this is a promising research direction. 

\subsection{Correlation between Probing and Downstream Tasks}
\label{sec:structure_infusion/analysis}

\begin{figure}[ht]
    \centering
    \includegraphics{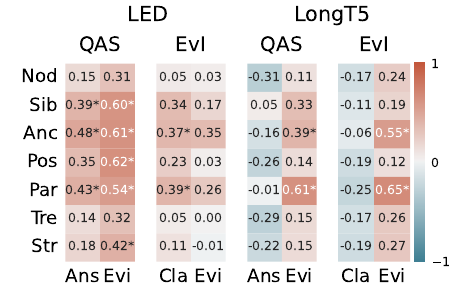}
    \caption{Pearson correlation between probing and downstream tasks. * denotes significance (p < 0.05).}
    \label{fig:correlation}
\end{figure}

To find associations between the representation of document structure and downstream task performance, we computed the Pearson correlation between probing and downstream task metrics \footnote{The absolute values from each set of bars in Fig.~\ref{fig:probing_structure_infusion} were paired with the unaggregated values from each column in Tab.~\ref{tab:downstream_tasks} for the same model.} (Fig.~\ref{fig:correlation}). All combinations of probing and downstream tasks for LED, and evidence selection and all probing tasks for LongT5 have a correlation greater or around 0. In contrast, the performance of LongT5 on QASPER answer generation and Evidence Inference classification is mostly negatively correlated with the probing task metrics. These were also the tasks with the least improvements through structure infusion. As they are decoder-based tasks, while evidence selection is encoder-based (§\ref{sec:appendix/implementation/downstream}), it seems that LongT5 has less need for structure infusion on decoder-based tasks.

For LED in both QASPER subtasks and Evidence Inference classification and for LongT5 in evidence selection on both Evidence Inference and QASPER, we see significant ($p < 0.05$)  correlation with the \texttt{ancestor} and \texttt{parent predecessor} probes, which measure the representation of relations between nodes on one directed path of \texttt{parent} edges. These usually have more defined semantic relationships among each other compared to nodes from different paths, e.g. a section heading has more relevant information about the paragraphs belonging to that section than about those in other sections. Our results suggest that better representation of these relations is associated with better downstream performance. 

\section{Conclusion}

In this work, we provided an in-depth analysis of the representation of abstract document structure in long-document Transformers. Experiments with our novel probing suite show that LED and LongT5 have learned to represent node function and hierarchical organization through pre-training without explicit supervision, with room for improvement.

To investigate the effect of infusing the aspects of document structure that are missing in Transformer inputs due to linearization, we developed a modular structure infusion framework. Probing shows that structure infusion enhances the internal representation of document structure, and we see performance improvements from structure infusion on QASPER and Evidence Inference, two downstream tasks where this has not been shown before. The significant correlation between several probing and downstream tasks suggests that it is indeed the improved representation of document structure that leads to downstream task performance gains.

Our probing, structure infusion and downstream task suite is easily extensible with new probing and downstream tasks and new types of infused information. While this work provides proof of the utility of our graph-based framework for documents from the scientific domain, the framework can be applied to other document types (e.g. web pages or conversation threads). Given that the addition of separator tokens between document elements can already increase performance, we deem applying our methods to documents with less well-defined structure promising. Our probing methods are fully compatible with the current generation of Transformer-based LLMs \cite{workshop2023bloom, touvron2023llama}, as long as the internal states of the model can be accessed. We hope that our contributions pave the path towards systematic study of the role of document structure in NLP.

\section*{Ethical Considerations}

Long documents lie at the core of text work, and structure is omnipresent in long documents. We believe that developing a better understanding of the role of document structure in NLP would allow us to build more efficient, robust, and interpretable systems for the analysis of long texts. We envision a trade-off between structural modeling capabilities of NLP systems (which, as we show, can be enhanced by providing explicit document structure) and the computational and storage overhead associated with processing additional structural information in the documents. Future work would investigate this trade-off and determine in which cases this overhead is justified. As document structure is openly present in documents and easily accessible by humans, we do not envision additional ethical risks or misuse scenarios due to the use of document structure in NLP modeling. Our work only uses data published under permissive licenses; our adaptations of this data are made available under permissive conditions as well.

\section*{Limitations}
We see our work as an important step towards the general study of the role of document structure in NLP modeling. Below we outline the limitations of our work, which present excellent opportunities for follow-up research.

\textbf{Dataset diversity.} Our work unifies structured document data from multiple sources. Yet all of this data originates form the scientific domain. There are several benefits to this: scientific documents are long, clearly licensed, and exhibit structure -- and the scientific domain offers multiple long-document processing tasks. In addition, focusing on one general domain allows us to control for domain shift during our measurements. We note that no part of our methodology is tailored to the particularities of the scientific domain -- and as long as source documents can be converted into the domain-agnostic ITG formalism, our methods should be easily adaptable to other domains like Wikipedia or conversation threads. Similarly, we limit our studies to the English language, as other languages face scarcity both in terms of available long-document Transformer models and academic texts. As more data and models become available, it will become possible to evaluate our findings in new contexts.

\textbf{Models and Tasks.} Our setup involves multiple probing tasks coupled with a range of structure infusion methods, resulting in a wide experimental grid. To make in-depth analysis feasible, we had to limit our focus on a few models and tasks. We chose two datasets which combine generative question answering, segment classification and document classification. Our experiments show that structure infusion can be useful for all tasks and models considered. This suggests that experiments on other tasks are a promising direction for future research, which is facilitated by our open implementation.

\textbf{Large language models.} While it would be technically possible to apply our kit to the recent decoder-only models such as LLaMA \cite{touvron2023llama} or BLOOM \cite{bigscience-2022-bigscience}, this would require substantial computational resources -- which illustrates the challenges of long-document processing by modern NLP models and does not constitute a limitation of our proposed approach. Similarly, commercially hosted models with increased input length such as GPT-4\footnote{\url{https://openai.com/gpt-4}} (32k tokens) and Claude \footnote{\url{https://www.anthropic.com/product}} (100k tokens) could be evaluated and infused with document structure -- yet their closed-source nature and lack of access to model weights prevents such investigation. We hope that the progress in efficient NLP and the ongoing open-source LLM development make such studies possible in the near future. This would also pave the way for investigating the effect of abstract document structure in zero-shot experiments.

\textbf{Other types of document structure} As noted in the "Document structure" paragraph in §\ref{sec:background}, we focus on investigating abstract document structure. We also mention visual and discourse structure as important structural properties of documents. While we don't study them here, this is done in current works, e.g. Huang et al. (\citeyear{10.1145/3503161.3548112}) or Du et al. (\citeyear{du-etal-2023-structure}). We believe that joint investigations of the different aspects of document structure are a promising direction for future work.

\textbf{Correlated model states.} The structure-infused models in this work were first pre-trained using a language modeling loss on probing or downstream task data, and then further fine-tuned using a task-specific loss.
The probing and downstream task datasets in our work are \emph{not identical}; thus, strictly speaking, the scores used to compute the correlation in Fig.~\ref{fig:correlation} come from models with the same structure infusion configuration, but not the same \emph{state}. We believe this to be unproblematic and expect the states to be comparable, since each model is pre-trained under the same regime. To confirm this, future work could create probing datasets from downstream task datasets to use the same model state in probing and downstream tasks -- at the cost of a drastic increase in the number of probing experiments. This technical limitation only pertains to §\ref{sec:structure_infusion/analysis} and Fig.~\ref{fig:correlation} and leaves all other results unaffected.

\section*{Acknowledgements}
 Funded by the European Union (ERC, InterText, 101054961). Views and opinions expressed are however those of the author(s) only and do not necessarily reflect those of the European Union or the European Research Council. Neither the European Union nor the granting authority can be held responsible for them. 

 We thank the anonymous reviewers for their helpful comments and suggestions.

\bibliography{anthology,custom}

\appendix

\section{Table of Probing Results}

See Tab.~\ref{tab:probing}

\begin{table*}[]
\centering
\begin{tabular}{lrrrrrrr}
\toprule
  & Nod & Sib & Anc & Pos & Par & Tre & Str \\
\midrule
LED & 93.98 & 64.93 & 89.53 & 86.05 & 85.68 & 84.12 & 41.49 \\
LED Atom & 92.75 & 60.26 & 87.30 & 65.53 & 84.82 & 82.41 & 40.64 \\
LED Rand & 88.21 & 58.36 & 86.73 & 56.44 & 82.90 & 73.76 & 35.33 \\
\midrule
tok-boundaries & 94.15 & 65.87 & 89.80 & 87.13 & 86.30 & 85.64 & 40.68 \\
tok-depth & 99.78 & 67.41 & 90.99 & 89.59 & 87.64 & \textbf{99.96} & 51.22 \\
tok-type & 95.39 & 66.70 & 90.23 & 88.64 & 87.12 & 87.06 & 42.16 \\
emb-depth & 99.90 & 68.55 & 90.21 & 94.09 & 87.83 & \textbf{99.96} & 54.54 \\
emb-type & 95.60 & 67.99 & 90.49 & 92.37 & 86.99 & 89.32 & 46.48 \\
emb-depth-tok-type & \underline{99.98} & \textbf{69.71} & 91.31 & 94.85 & \textbf{88.85} & \textbf{99.96} & \underline{55.87} \\
emb-type-tok-type & 95.54 & 69.34 & 90.74 & 92.30 & 88.23 & 90.26 & 46.14 \\
emb-type-tok-depth & \textbf{100.00} & \underline{69.57} & \underline{91.72} & \underline{95.97} & 88.31 & \textbf{99.96} & 54.43 \\
emb-depth-tok-depth & 99.95 & 69.43 & \textbf{91.81} & \textbf{96.30} & \underline{88.68} & \textbf{99.96} & \textbf{55.94}\\
\midrule \midrule
LongT5 & 95.28 & 65.85 & 89.38 & 91.95 & 86.13 & 87.88 & 42.97 \\
LongT5 Atom & 91.84 & 50.79 & 86.60 & 61.05 & 83.77 & 78.90 & 34.68 \\
LongT5 Rand & 88.21 & 57.41 & 84.81 & 57.97 & 81.54 & 73.40 & 33.49 \\
\midrule
tok-sep & 95.88 & 66.93 & 90.41 & 93.16 & 87.62 & 88.76 & 45.47 \\
tok-depth & 99.90 & 67.79 & 91.20 & 95.82 & \textbf{88.45} & \textbf{99.96} & 52.51 \\
tok-type & 95.99 & 67.96 & 90.92 & 94.80 & 87.59 & 89.26 & 44.60 \\
emb-depth & 99.92 & 67.75 & 90.94 & 98.32 & 87.45 & \textbf{99.96} & 51.92 \\
emb-type & 95.85 & 68.23 & 90.33 & 96.13 & 86.79 & 89.92 & 45.89 \\
emb-depth-tok-type & \textbf{99.98} & 67.88 & 90.52 & \textbf{98.86} & \underline{88.25} & \textbf{99.96} & \underline{54.09} \\
emb-type-tok-type & 96.07 & \underline{68.30} & 90.85 & 96.75 & 87.44 & 91.13 & 46.73 \\
emb-type-tok-depth & \textbf{99.98} & 67.99 & \textbf{91.53} & 97.98 & 87.92 & 99.74 & 49.07 \\
emb-depth-tok-depth & \underline{99.97} & \textbf{68.66} & \underline{91.27} & \underline{98.70} & 87.15 & \textbf{99.96} & \textbf{54.40} \\
\bottomrule
\end{tabular}
\caption{Probing result numbers for Fig. \ref{fig:probing_structure_infusion} and from Tab.~\ref{tab:probing_baselines} for comparison. The best result per model is printed in bold, the second best is underlined.}
\label{tab:probing}
\end{table*}

\section{Implementation Details}
\label{sec:appendix/implementation}

\subsection{Models}
\label{sec:appendix/implementation/models}
In all experiments, we used the huggingface Transformers\footnote{\url{https://huggingface.co/}} \cite{wolf-etal-2020-Transformers} implementations and weights of LED base (162M parameters, \citealt{beltagy2020longformer}) and LongT5 base with transient global attention (220M parameters, \citealt{guo-etal-2022-longt5}).

\subsection{Probing}
\label{sec:appendix/implementation/probing}

\paragraph{Dataset.}

\begin{table}
    \centering
    \begin{tabular}{llrrr}
\toprule
 & Label & Dev & Test & Train \\
\midrule
\multirow[t]{3}{*}{\texttt{Anc}} & False & 7665 & 7999 & 23488 \\
 & True & 7665 & 7999 & 23488 \\
 & \hspace{1.5mm}\textbf{Total} & 15330 & 15998 & 46976 \\
\cline{1-5}
\multirow[t]{4}{*}{\texttt{Nod}} & Paragraph & 2353 & 2369 & 7046 \\
 & Section & 2278 & 2298 & 6708 \\
 & Subsection & 1250 & 1262 & 3611 \\
 & \hspace{1.5mm}\textbf{Total} & 5881 & 5929 & 17365 \\
\cline{1-5}
\multirow[t]{3}{*}{\texttt{Par}} & False & 7665 & 7999 & 23488 \\
 & True & 7665 & 7999 & 23488 \\
 & \hspace{1.5mm}\textbf{Total} & 15330 & 15998 & 46976 \\
\cline{1-5}
\multirow[t]{4}{*}{\texttt{Pos}} & Begin & 3049 & 3180 & 9406 \\
 & End & 3049 & 3180 & 9406 \\
 & Inside & 3049 & 3180 & 9406 \\
 & \hspace{1.5mm}\textbf{Total} & 9147 & 9540 & 28218 \\
\cline{1-5}
\multirow[t]{3}{*}{\texttt{Sib}} & False & 7665 & 7999 & 23488 \\
 & True & 7665 & 7999 & 23488 \\
 & \hspace{1.5mm}\textbf{Total} & 15330 & 15998 & 46976 \\
\cline{1-5}
\multirow[t]{9}{*}{\texttt{Str}} & 1 & 2939 & 3044 & 8946 \\
 & 2 & 2939 & 3044 & 8946 \\
 & 3 & 2939 & 3044 & 8946 \\
 & 4 & 2912 & 3018 & 8823 \\
 & 5 & 1840 & 1926 & 5560 \\
 & 6 & 985 & 1124 & 3161 \\
 & 7 & - & 10 & 5 \\
 & 8 & - & - & 5 \\
 & \hspace{1.5mm}\textbf{Total} & 14554 & 15210 & 44392 \\
\cline{1-5}
\multirow[t]{6}{*}{\texttt{Tre}} & 1 & 2892 & 2895 & 8642 \\
 & 2 & 2892 & 2895 & 8642 \\
 & 3 & 1634 & 1639 & 4872 \\
 & 4 & - & 3 & 1 \\
 & 5 & - & - & 1 \\
 & \hspace{1.5mm}\textbf{Total} & 7418 & 7432 & 22158 \\
\bottomrule
\end{tabular}

    \caption{Label distribution across probing tasks. \texttt{Anc}: \texttt{Ancestor}; \texttt{Nod}: \texttt{Node type}; \texttt{Par}: \texttt{Parent predecessor}; \texttt{Pos}: \texttt{Position}; \texttt{Sib}: \texttt{Sibling}; \texttt{Str}: \texttt{Structural}; \texttt{Tre}: \texttt{Tree depth}.}
    \label{tab:dataset}
\end{table}

Our probing dataset is split 0.6/0.2/0.2 across train, dev, and test using in-document balancing. For boolean and the \texttt{position} probe we see a uniform distribution of instances per label, compared to the \texttt{node type} probe where subsections occur not in all documents, resulting in a non-uniform distribution. The \texttt{structural} and \texttt{tree depth} probes naturally feature a diverse set of labels and instances. A full overview of the label distribution can be found in Tab.~\ref{tab:dataset}.

\begin{table}
\centering
\begin{tabular}{lc}
\hline
\multicolumn{2}{c}{Training}                                      \\ \hline
Batch size & 4 (VR), 64 (AT) \\
Epochs & 20 \\
Patience & 10 \\
\hline
\multicolumn{2}{c}{Optimization} \\ 
\hline    
Algorithm & Adam \cite{DBLP:journals/corr/KingmaB14} \\
$\beta_1, \beta_2$ & $0.9$, $0.999$ \\
$\epsilon$ & $10^{-8}$ \\
Weight decay & 0.01 \\
Learning rate & $10^{-3}$(LED), $10^{-1}$(LongT5) \\
\end{tabular}
\caption{Vanilla and random (VR), and atomic (AT) configuration hyperparameters.}
\label{tab:appendix/probing_hyperparams}
\end{table}

\paragraph{Implementation and hyperparamenters.}

Our probing kit is implemented using the AllenNLP library~\cite{gardner-etal-2018-allennlp}. We stack a frozen pre-trained Transformer model with an endpoint span extractor from AllenNLP, extracting and concatenating the first and last token of a given span. Our hyperparameters are described in Tab.~\ref{tab:appendix/probing_hyperparams}.

\begin{figure*}
    \centering
    \includegraphics{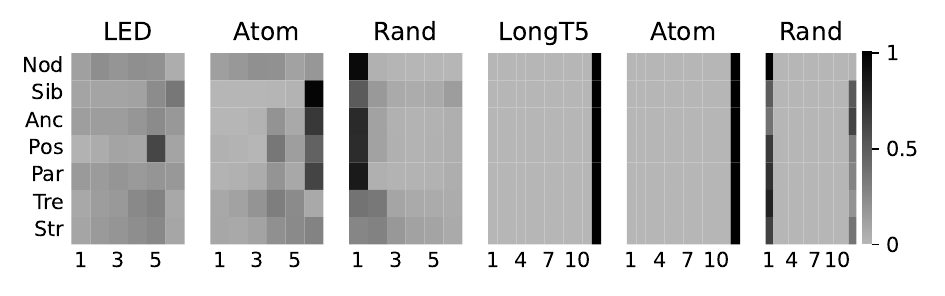}
    \caption{Layer utilization in probing of the vanilla LED and LongT5 models.}
    \label{fig:layer_util}
\end{figure*}

\paragraph{Layer utilization.}

The layer utilization shown in Fig.~\ref{fig:layer_util} reveals differences between the probed models and their controls. For LED, the vanilla configuration shows a more uniform layer utilization compared to the control configurations. The atomic control puts more weight on the last layer for all probes except \texttt{node type} and \texttt{tree depth}. For LongT5, both vanilla and atomic put all weight on the last layer. For LED and LongT5, the random control mostly uses the first layer, which has also been observed in other works~\cite{voita-titov-2020-information}. The random control relies solely on the input embeddings, as there is no additional information in the Transformer layers. Input words such as "Introduction" and the number of tokens in a text node can be used to infer the node type. Node type and word overlaps between two nodes can give hints to the relation between two nodes. With LongT5, the intermediate layers are not used at all. 

As the atomic control cannot compare the position embeddings of different nodes, it makes full use of the contextualization through the entire forward pass. To solve the \texttt{node type} task, the length of a node provides useful information. It is retained in the atomic position embeddings, explaining the more uniform layer utilization on this probe. The random control puts most weight on the the first layer, which has also been observed in other works~\cite{voita-titov-2020-information}. It relies on the input embeddings, as there is no additional information in the Transformer layers.

\subsection{Structure Infusion}
\label{sec:appendix/implementation/infusion}
\paragraph{Embeddings.} Structural embeddings are added to the token embeddings of each token in a node (including special tokens) before the first encoder self-attention layer (Fig.~\ref{fig:structure_infusion}). They were initialized according to a Gaussian distribution with mean 0 and standard deviation 0.0305 (LED) and 4.875 (LongT5). Standard deviation for LED was chosen to be the same as the standard deviation of the absolute linear position embeddings matrix. As LongT5 does not have absolute position embeddings, the standard deviation for structural embedding initialization was chosen to result in the same ratio of token embedding standard deviation to structural embedding standard deviation as for LED. 

\paragraph{Special tokens.} Special tokens are prepended to the tokens of the respective node, leading to an increase in total sequence length (Fig.~\ref{fig:structure_infusion}). They were initialized using the \texttt{resize\_token\_embeddings()} function in the model implementation.

\paragraph{Number of added parameters.} 
\begin{table}
\centering
\begin{tabular}{lc}
\hline
Config                              & $n_{parameters}$ \\ \hline
\texttt{tok-type}  & 3K               \\
\texttt{emb-type}  & 3K               \\
\texttt{tok-depth} & 15K              \\
\texttt{emb-depth} & 15K              \\ \hline
\end{tabular}
\caption{Number of added parameters in structure infusion}
\label{tab:appendix/infusion_parameters}
\end{table}

For the number of added parameters for each infusion configuration see Tab.~\ref{tab:appendix/hyperparams}. Each special token and each embedding adds $d_{model}$ parameters to a model ($d_{LED} = d_{LongT5} = 768$). There were 4 structural tokens / embeddings and 20 node depth tokens / embeddings.

\subsection{Pre-Training}
\label{sec:appendix/implementation/pretraining}

\begin{table}
\centering
\begin{tabular}{lc}
\hline
\multicolumn{2}{c}{Masking}                                                            \\ \hline
Noise density                & 3\%                                                     \\
Mean noise span length       & {[}4,8,12{]}*                                           \\ \hline
\multicolumn{2}{c}{Training}                                                           \\ \hline
Batch size                   & 16 (PT), 8 (FT)                                                     \\
Steps            & 15000 (PT)                                              \\
 & 10200 (FT) \\ \hline
\multicolumn{2}{c}{Optimization} \\  \hline    
Algorithm & AdamW [1]                                         \\
$\beta_1, \beta_2$ & $0.9$, $0.999$                                          \\
$\epsilon$         & $10^{-8}$                                               \\
Weight decay       & 0.01                                                    \\
Learning rate                & $10^{-5}$ (LED) \\
 & $10^{-4}$ (LongT5) \\
Warmup & Linear (PT), - (FT) \\
Warmup steps & 500 (PT), - (FT)
\end{tabular}
\caption{Pre-training (PT) and fine-tuning (FT) hyperparameters. *: Mean noise span length is chosen uniformly from the given values for each input sequence. [1] \citealt{DBLP:conf/iclr/LoshchilovH19}}
\label{tab:appendix/hyperparams}
\end{table}

All structure infused models and baselines were pre-trained on the respective probing or evaluation dataset using a "T5-style" denoising task. Noise was added to the model input using code provided by the authors of the T5 \cite{10.5555/3455716.3455856} paper\footnote{\url{https://github.com/google-research/text-to-text-transfer-transformer}}, which replaces spans of tokens in the input with numbered mask tokens. The mask tokens were initialized using the \texttt{resize\_token\_embeddings()} function in the model implementation. Masking is controlled by two hyperparameters: \textit{noise density}, the proportion of masked tokens in the input, and \textit{mean noise span length}. We chose the noise density as 3\%, the mean noise span length was uniformly chosen for each input sequence from 4, 8 or 12 tokens.

The model is trained with a cross entropy loss to generate each mask token followed by the tokens replaced by that mask, respecting the order of masked spans. To save computation, only one checkpoint was pre-trained for each combination of model, infusion configuration and dataset. This checkpoint was used in all replicates of a downstream experiment.

\paragraph{Training hyperparameters}
For training hyperparameters, see Tab.~\ref{tab:appendix/hyperparams}. 

The only optimized hyperparameter is the learning rate, which was done by grid search with the respective non-pretrained vanilla configuration on the QASPER dataset.

\

\subsection{Downstream Tasks}
\label{sec:appendix/implementation/downstream}

\subsubsection{QASPER} 
\label{sec:appendix/implementation/downstream/QASPER}

\paragraph{Dataset conversion.} Each entry in the QASPER dataset \cite{dasigi-etal-2021-dataset} consists of a paper title, abstract, full text in the form of a list of sections with section name and corresponding paragraphs, a list of figures and tables, as well as a list of questions, answers and evidence. We converted the QASPER dataset into the Intertext Graph (ITG) format \cite{10.1162/coli_a_00455} creating a node for the title, abstract, each section title and each paragraph, as well as figures and tables. We added an additional \texttt{abstract} node with the content "Abstract" to serve as the parent for the abstract text. 

All answer types (extractive, abstractive, yes/no, unanswerable) were mapped to a single reference answer string for each question as done by the dataset authors. The provided evidence strings were mapped to the ITG nodes through string matching, which which was successful for 99.35\% of evidence pieces from the original dataset. For 0.41\%, there was no match, and for 0.24\% there were multiple matches, which were discarded. Questions, answers and evidence are stored in the ITG metadata. We follow the original data splits, resulting in 888 train, 281 validation and 416 test documents.

\paragraph{Model input.} For LED, model input was formed as "\texttt{<s> [question] </s> [document]}". For LongT5, the initial \texttt{<s>} token was not used, as it is not pre-trained with this token. Figures and tables were discarded for model input.

\paragraph{Evaluation.} QASPER evaluation was implemented by adapting the evaluation script provided by the creators of the dataset\footnote{\url{https://github.com/allenai/qasper-led-baseline}}. If there are multiple reference answers to a question, the answer that results in the highest score is chosen as the gold standard. Answer generation is evaluated with a token-level F1 score as in SQuAD \cite{rajpurkar-etal-2016-squad}. Evidence selection is evaluated with a node-level F1 score.

\paragraph{Answer generation.} Answers were generated with beam search, using 4 beams, length penalty 1.0 and a maximum generated length of 100 tokens.

\paragraph{Evidence selection.} Evidence selection was implemented as paragraph classification. There can be multiple evidence paragraphs for a question. The final encoder hidden state $h$ of the first token of each \texttt{paragraph} node in a document is used as the representation for the paragraph. This vector is passed through a fully connected linear layer $W_1$ followed by a tanh nonlinearity  and a linear layer $W_2$ projecting to the score vector $s \in \mathbb{R}^2$ for evidence and no-evidence. 
\begin{equation}
    s = W_2 \tanh(W_1h), \; W_1 \in \mathbb{R} ^{d \times d}, \; W_2 \in \mathbb{R}^{d \times 2}
\end{equation}

\paragraph{Fine-tuning.} Models pre-trained as described above on the QASPER train documents were fine-tuned on  with the hyperparameters given in Tab.~\ref{tab:appendix/hyperparams}. Answer generation and evidence selection were trained with cross entropy loss: \begin{equation}
    \mathcal{L}=w_A\mathcal{L}_{Answer} + w_E\mathcal{L}_{Evidence}
\end{equation}
For LED and LongT5 the loss weights were set to $w_A=w_E=0.5$. The checkpoint with the best score on the dev set was used for evaluation.

\subsubsection{Evidence Inference}

\paragraph{Dataset conversion.} Evidence Inference 2.0 \cite{deyoung-etal-2020-evidence} is provided as sets of articles, prompts and labels with evidence. The article full texts are provided as plain text files and NXML files following the PubMed DTD schema\footnote{\url{https://pubmed.ncbi.nlm.nih.gov/download/}}. We used the parser from the dataset creators\footnote{\url{https://github.com/jayded/evidence-inference}} to parse the NXML files, and converted the output to the ITG format. We added an additional \texttt{abstract} node with the content "Abstract" to serve as the parent for the abstract text.

Evidence annotations are given as character offsets pertaining to the articles in plain text format. We transform this span selection problem to a node classification problem by mapping evidence strings to ITG nodes. Evidence text at a given offset is extracted from a text file and then matched against ITG nodes using fuzzysearch\footnote{\url{https://github.com/taleinat/fuzzysearch}}. Full string matching resulted in low recall, because of small differences between the plain text files and NXML files. For 92.03\% of evidence spans, we find exactly one ITG node, for 5.10\% we find no node, and for 2.07\% we find more than one node, which are discarded. The prompts, labels and evidence for a document are stored in the ITG metadata. We follow the original data splits, resulting in 3562 train, 443 validation and 449 test documents.

\paragraph{Model input.} For LED, model input was formed as "\texttt{<s> With respect to [outcome], characterize the reported difference between patients receiving [intervention] and those receiving [comparator]. </s> [document]}". For LongT5, the initial \texttt{<s>} token was not used, as it is not pre-trained with this token.

\paragraph{Evaluation.} Evidence Inference classification is evaluated with macro F1 score. Evidence selection  is evaluated with a node-level F1 score. If there are multiple annotations to a prompt, the annotation that results in the highest score is chosen. We chose to implement the evaluation similar to QASPER evaluation for consistency, and thus different from the implementation by the creators of the dataset. The main differences are (1) the conversion of evidence selection to a node classification task and (2) choosing the classification annotation that results in the highest score, where in the original implementation the class with the highest number of annotations is chosen as the gold standard. 

\paragraph{Classification.} To get the class of a prompt-document pair, a vector representation $v$ of the document is passed through a fully connected layer $M_1$, followed by a tanh nonlinearity and a linear layer $M_2$ projecting to the score vector $l \in \mathbb{R}$.
\begin{equation}
    l = M_2(\tanh(M_1(v))), \; M_1 \in \mathbb{R}^{d \times d}, \; M_2 \in \mathbb{R}^{d \times 3}
\end{equation}
For LED, $v$ was chosen as the final encoder hidden state of the initial \texttt{<s>} token, because it has global attention. As LongT5 does not have configurable global attention, a dummy \texttt{</s>} token was input to the decoder, which has full cross attention over the input document. The final decoder hidden state of this token served as $v$ for LongT5. 

\paragraph{Evidence selection.} Evidence selection was implemented as for QASPER (§\ref{sec:appendix/implementation/downstream/QASPER}).

\paragraph{Fine-tuning.} Models pre-trained as described above on the Evidence Inference train documents were fine-tuned with the hyperparameters given in Tab~\ref{tab:appendix/hyperparams}. Classification and evidence selection were trained with cross entropy loss:
\begin{equation}
    \mathcal{L} = w_C\mathcal{L}_{Classification} + w_E\mathcal{L}_{Evidence}
\end{equation}
For LED, the loss weights were set to $w_C=w_E=0.5$. For LongT5, they were set to $w_C=0.25, \; w_E=0.75$. The checkpoint with the best score on the dev set was used for evaluation.

\subsection{Computation}
\label{sec:appendix/implementation/computation}
Experiments were performed on NVIDIA A100, A180 and A6000 GPUs. Depending on the GPU size and speed, pre-training, probing (all 7 tasks) and downstream task experiments took $~$1-2 days. Estimating an average of 1.5 days per experiment, the total number of GPU days is 264 (26 probing runs, 30 pre-training runs, 120 downstream fine-tuning runs).

\subsection{Use of AI Assistants in Development}

Some of the code for the structure infusion framework was developed with assistance from GitHub Copilot\footnote{\url{https://github.com/features/copilot}}.

\end{document}